\documentclass[11pt,a4paper]{article}
\usepackage[hyperref]{acl2021}
\usepackage{times}
\usepackage{latexsym}

\usepackage{microtype}
\usepackage{algorithm}
\usepackage{algorithmic}

\usepackage{multirow}
\usepackage{hhline}
\usepackage{physics}
\usepackage{graphicx}
\usepackage{tikz}
\usepackage{amsfonts}
\usepackage{tabularx}
\usepackage[normalem]{ulem}
\usepackage{booktabs}
\usepackage{trackchanges}
\usepackage{ulem}
\usepackage{xcolor}
\addeditor{YJ}

\aclfinalcopy 

\title{Simple Text Detoxification by Identifying a Linear Toxic Subspace in Language Model Embeddings}

\author{Andrew Wang \\
  University of Virginia \\
  \texttt{ajw7uhj@virginia.edu} \\\And
  Mohit Sudhakar \\
  University of Virginia \\
  \texttt{ms5sw@virginia.edu} \\\And
  Yangfeng Ji \\
  University of Virginia \\
  \texttt{yangfeng@virginia.edu} \\}

\date{}

\begin{document}
\maketitle
\begin{abstract}
  Large pre-trained language models are often trained on large volumes of internet data, some of which may contain toxic or abusive language.
  Consequently, language models encode toxic information, which makes the real-world usage of these language models limited.
  Current methods aim to prevent toxic features from appearing generated text.
  We hypothesize the existence of a low-dimensional toxic subspace in the latent space of pre-trained language models, the existence of which suggests that toxic features follow some underlying pattern and are thus removable.
  To construct this toxic subspace, we propose a method to generalize toxic directions in the latent space. We also provide a methodology for constructing parallel datasets using a context based word masking system. Through our experiments, we show that when the toxic subspace is removed from a set of sentence representations, almost no toxic representations remain in the result.
  We demonstrate empirically that the subspace found using our method generalizes to multiple toxicity corpora, indicating the existence of a low-dimensional toxic subspace. 
\end{abstract}

\section{Introduction}
Transformer language models have recently found great success in tasks such as text classification, owing in large part to the large amounts of data used to train the model \citep{Bert}. Much of this data is derived from online sources without much vetting, and a major consequence is that training data contains toxic language \citep{RealTox}. As a result, language models learn toxic behavior \citep{DBLP:journals/corr/abs-1908-07125}. Removing toxic text directly from the training data has proven difficult due to the amount of the training data present \citep{DBLP:journals/corr/abs-2004-13637}, therefore recent works consider post-hoc methods to remove toxic information from pre-trained language models. 

Research along this line generally aims to prevent toxic features from appearing in language model output. For example, a number of approaches treat toxicity as an attribute that can be separated from the primary meaning of sentence. In controlled text generation, progress has been made in removing toxic behavior while maximizing fluency \citep{DBLP:journals/corr/abs-1912-02164}. In style transfer, the meaning of a toxic sentence is mapped onto a non-toxic target sentence \cite{DBLP:journals/corr/abs-2102-05456}. 

We hypothesize that instances of toxicity lie along some lower dimensional subspace within pretrained lanugage model embeddings. For sentence level toxicity removal, toxicity is predicated on the existence of context-dependent toxic features. For instance, consider the toxic examples in Table \ref{tabel:ex}. In each case, the toxicity manifests in a fraction of the total vocabulary present within the sentence. We are interested in whether the existence of these toxic features follow some predictable pattern. To identify this subspace, we will use principal component analysis to extract toxic features. Using the toxic subspace, we can very efficiently detoxify a sentence embedding. We thus propose a novel yet simple way to perform post-hoc toxicity mitigation. 

We evaluate how well our toxic subspace can explain toxicity via sentence-level toxicity classification tasks. If we remove the toxic subspace from sentence embeddings, then we should observe a corresponding decrease in model accuracy. We will not include as part of our toxicity removal evaluations any decoding tasks. The decoding step introduces variables that interfere with our evaluation of the subspace. We would be unable to ascertain whether detoxification occurred because we removed the subspace or because of the decoding step.


We distinguish our task from other related tasks that also manipulate the latent space such as removing undesired biases in language models \citep{DBLP:journals/corr/BolukbasiCZSK16a,DBLP:journals/corr/abs-1904-03035,liang-etal-2020-towards} or disassociating protected groups from toxic contexts \citep{46743,park-etal-2018-reducing,DBLP:journals/corr/abs-2102-00086}. Debiasing tasks generally involve manipulating the latent space such that bias classes such as different genders are equally represented, or they involve relabelling the dataset such that protected groups are not disproportionately present in toxic examples. Our task instead aims to extract toxic features such that we can mitigate toxicity.

We provide a method to identify a toxic subspace in the embeddings of language models such as BERT. More specifically we contribute, as components of our overall subspace identification method, 1) an algorithm to parallelize non-parallel toxicity data, and 2) a set of heuristics for identifying the eigenvectors of the toxic subspace. We observe that our method can successfully identify a subspace that causes toxicity. 

\section{Related Work}

Probings of pretrained language models such as BERT and GPT-2 find that these language models associate specific contexts with toxic language \citep{nozza-etal-2021-honest,ousidhoum-etal-2021-probing}.
These probing techniques create a prompt in the format of ``a [\textit{protected group}] [\textit{does a specific action}] because they..." where different protected groups and actions can be used in the prompt.
A language model such as GPT-2 must then complete the prompt, after which, the continuations are analyzed.
This technique resembles the problem of toxic degeneration \cite{RealTox}, wherein a non-toxic prompt may still induce a language model to generate toxic text.
For instance, within Real Toxicity Prompts, a dataset of 100k sentences each split in half into a prompt and continuation, toxicity was generally found to be confined to either the prompt \textit{or} the continuation, not both, suggesting that toxic continuations can result from non-toxic prompts \cite{RealTox}.

Furthermore, if toxicity is generally not present in either the prompt-half of a toxic sentence or the continuation-half of a toxic sentence, this implies that toxicity is concentrated in a smaller subset of words or phrases within that sentence.
Either toxicity is concentrated in sentences in a predictable manner, or certain contexts cause toxicity more than other contexts, suggesting that toxicity can be explained by a lower dimensional subspace in language model embeddings which can then be removed. 

Removing a subspace responsible for a certain attribute has been well studied in the context of language model debiasing \citep{DBLP:journals/corr/BolukbasiCZSK16a,DBLP:journals/corr/abs-1904-04047,liang-etal-2020-towards}.
These approaches remove bias by first identifying bias vectors for each sentence.
Principal component analysis is performed on these vectors to produce a bias subspace.
To remove bias from an input sentence, the sentence's bias component is identified using the bias subspace and removed.
A number of methods have been proposed to find the aforementioned bias vectors.
In particular, \citet{liang-etal-2020-towards} creates a parallel training corpus for male/female debiasing by swapping gender pronouns (eg. “she was a doctor” and “he was a doctor”), then computing the difference between each sentence embedding.
However, we cannot do this for toxicity for three reasons: (1) there are many offensive words, (2) a word’s offensiveness often depends on context, (3) offensive words do not have a corresponding ``unoffensive version.''
Thus, we propose our own method of creating a parallel corpus and removing toxic features.




\section{Methodology}
The intuition behind our approach is as follows. For any given toxic sentence, holding context constant, we analyze the differences in latent space when words that contribute to toxicity are removed. Consider the toxic sentences in Table \ref{tabel:ex}. In each example we remove the toxic words to create a masked sentence. We can extract a toxic ``direction" by feeding both toxic and masked sentences through a language model encoder such as BERT and finding the difference between the resulting sentence representations. Given many toxic directions, we find a subspace that best explains these directions. Thus, given $\vb{X}_1 = \{\vb{x}_1^{(1)}, \vb{x}_2^{(1)}, ..., \vb{x}_n^{(1)}\}$, a set of toxic sentences, and $\vb{X_{2}} = \{\vb{x}_1^{(2)}, \vb{x}_2^{(2)}, ..., \vb{x}_n^{(2)}\}$ a non-parallel set of non-toxic sentences, we find and remove a subspace $V_t$ in pretrained language model embeddings responsible for toxicity.

\begin{table}[tp]
\begin{tabularx}{\linewidth}{l|X}
\hline
\textbf{}                & \textbf{Sentence} \\ \hline
Toxic           & you're a complete idiot if that's what you think                  \\
\textit{Non-Toxic} & \textit{you're a complete [MASK] if that's what you think}                  \\ \hline
Toxic           & no stop pulling us into this crap                  \\
\textit{Non-Toxic} & \textit{no stop pulling us into this [MASK]}                  \\ \hline
Toxic           & but hillary's a liar                  \\
\textit{Non-Toxic} & \textit{but hillary's a [MASK]}                  \\ \hline
Toxic           & another stupid hateful story by a clueless writer                  \\
\textit{Non-Toxic} & \textit{another [MASK] [MASK] story by a clueless writer}                  \\ \hline
\end{tabularx}
\caption{\label{tabel:ex} Examples of toxic token masking using our token masking algorithm.}
\end{table}

In order to compile a set of toxic directions, we require the preparation of a parallel toxic and non-toxic training corpus. To produce this parallel corpus, we begin by training a binary classifier \textit{MaskerModel} on $\vb{X}_1$ and $\vb{X}_2$ to perform toxicity classification. Afterwards, for each sentence in $\vb{X}_1$, we use \textit{MaskerModel} to identify and mask the tokens that contribute to style. The result is a discrete, style-agnostic form $z(\vb{x}_i^{(1)})$. We thus define the set of newly created non-toxic sentences $\vb{X}_{nt}$.
\begin{equation*}
    \vb{X}_{nt} = \{z(\vb{x}_i^{(1)}) : \vb{x}_i^{(1)} \in \vb{X_1}\}
\end{equation*}
We then encode each toxic sentence and non-toxic sentence using \textit{Encoder}, a pretrained language model such as BERT with parameters $\theta$. We define the set of toxic embeddings $\vb{W}_t$ and the set of non-toxic embeddings $\vb{W}_{nt}$.
\begin{align*}
    \vb{W}_t &= \{\textit{Encoder}(\vb{x}_i^{(1)}; \theta) : \vb{x}_i^{(1)} \in \vb{X_1}\} \\
    \vb{W}_{nt} &= \{\textit{Encoder}(\vb{x}_i^{(nt)}; \theta) : \vb{x}_i^{(nt)} \in \vb{X}_{nt}\}
\end{align*}

With the creation of our parallel dataset, we extend the method developed by \citep{liang-etal-2020-towards} to identify and remove the toxic subspace. We compute the difference between each pair of toxic and non-toxic sentences. We define $\vb{d}_t$ as the direction of toxicity and $\vb{d}_{nt}$ as the direction non-toxicity. 
\begin{align*}
    \vb{d}_t &= \{\vb{w}_i^{(t)} - \vb{w}_i^{(nt)} : \vb{w}_i^{(t)} \in \vb{W_t}, \vb{w}_i^{(nt)} \in \vb{W}_{nt}\} \\
    \vb{d}_{nt} &= \{\vb{w}_i^{(nt)} - \vb{w}_i^{(t)} : \vb{w}_i^{(t)} \in \vb{W_t}, \vb{w}_i^{(nt)} \in \vb{W}_{nt}\}
\end{align*}
These direction vectors reveal the transformation necessary to make non-toxic sentences toxic and toxic sentences non-toxic. We perform principal component analysis on $\vb{D} = \vb{d}_t \cup \vb{d}_{nt}$ and select the top $k$ eigenvectors to constitute the subspace $V$ responsible for both toxicity and non-toxicity. We then apply a set of heuristics to isolate the set of eigenvectors responsible specifically for toxic subspace $V_t$. We define these heuristics further below.  

Here, we distinguish our work from traditional biases such as gender bias. We define toxic bias as an assumption that the language model perpetuates toxic language due to training data. In gender bias removal, the aim is to find an intermediate representation in between gender classes such as male or female in the latent space \citep{DBLP:journals/corr/BolukbasiCZSK16a,liang-etal-2020-towards}. On the contrary, since our definition of toxic bias implies two classes, toxic and non-toxic, finding an intermediate representation between those two classes will not debias the model. We require toxic sentences to become completely non-toxic, instead of some half-toxic intermediate representation between toxic and non-toxic classes. In other words, our task differs from previous tasks in that, in principle, toxicity should not exist as a bias class in the first place.

We call to attention the difference between what we call a toxic bias and bias that occurs when performing toxicity classification. While a toxic bias is a predisposition towards toxic language due to training data, bias resulting from toxicity classification usually takes the form of incorrectly classifying as toxic a sentence mentioning protected groups \citep{46743} or containing speech patterns associated with these groups \citep{sap-etal-2019-risk} because of co-occurrence in the dataset. We highlight this distinction because the latter form of bias is often also referred to as toxic bias in other literature.

\subsection{Parallel Corpus Creation}
For each toxic sentence, we require an equivalent non-toxic sentence in order to compute a toxic direction. We previously called to attention the fact that, in a toxic sentence, a small subset of tokens disproportionately contribute to toxicity. Since we ultimately aim to explain the occurrence of these toxic tokens, our method of making a toxic sentence non-toxic should focus on these toxic tokens. 

Given a binary classifier \textit{MaskerModel} and a set of toxic sentences $\vb{X}_1$, we use \textit{MaskerModel} to systematically mask toxic tokens for each toxic sentence $\vb{x}_i^{(1)} \in \vb{X_1}$ until $\vb{x}_i^{(1)}$ becomes non-toxic. Each $\vb{x}_i^{(1)}$ contains $n$ tokens, such that $\vb{x}_i^{(1)} = \{t_1, t_2, ..., t_n\}$. Intuitively, we find and mask the token $t_s \in \vb{x}_i^{(1)}$ that contributes the most to toxicity. To find $t_s$, we define an auxiliary function \textit{Mask}$(\vb{x}_i, j)$ that masks the $j^{th}$ token of $\vb{x}_i^{(1)}$. For instance, given a sentence $s$ to be ``The quick brown fox jumps," \textit{Mask}($s$, $1$) returns ``[MASK] quick brown fox jumps," and \textit{Mask}($s$, $2$) returns ``The [MASK] brown fox jumps." For a sentence $\vb{x}_i^{(1)}$, we can construct $n$ such masked sequences. We define $\vb{Z}$ as a set of possible masked sequences
\begin{equation*}
    \vb{Z} = \{\textit{Mask}(\vb{x}_i^{(1)}, j) : j \in \mathbb{N}, j \leq n\} 
\end{equation*}

For each masked sentence $\vb{z}_i \in \vb{Z}$, we feed $\vb{z}_i$ through \textit{MaskerModel}. We determine $\vb{z}_{i,s}$, where $\vb{z}_{i,s}$ is the masked sentence with the smallest probability of being toxic according to \textit{MaskerModel}. An alternate way of considering $\vb{z}_{i,s}$ is as the masked sentence with token $t_s$ already masked. We can repeat our procedure on $\vb{z}_{i,s}$ to obtain a new $\vb{z}_{i,s}$ with two masked tokens, repeat the process again to obtain a $\vb{z}_{i,s}$ with three masked tokens, so on and so forth until the probability of $\vb{z}_{i,s}$ being toxic has lowered beneath a user-defined threshold, which we set to $.25$. We make the greedy assumption that minimizing the probability of being toxic at each iteration minimizes the probability that the final output $\vb{z}_{i,s}$ is toxic.

Our approach eliminates the need for any explicit set of predefined style-indicator tokens. Instead, we use context to predict which tokens are the most stylistically significant. Moreover, we provide a discrete, clear-box approach towards style neutralization, in which every masking decision can be examined and scrutinized. 

Masking tokens in excess will cause the semantic meaning of $\vb{z}_{i,s}$ to deviate significantly from the semantic meaning of its original sentence $\vb{x}_i^{(1)}$. Therefore, after each iteration, we compute the cosine similarity between the modified sequence and the original toxic input sequence, discarding the current observation if the cosine similarity begins to decay.

\subsection{Subspace Removal}
\begin{figure}
    \centering
    \begin{tikzpicture}
    \draw[thick,->,>=latex] (0,0) -- (2,4) node[midway, above left] {$\vb{w}_t$};
    \draw[thick,->] (0,0) -- (3,0) node[above] {$V_t$};
    \draw[thick,->,>=latex,red] (0,0) -- (2,0) node[midway, below] {$\vb{d}_t$};
    \draw[thick,->,>=latex,black!60!green] (2,0) -- (2,4) node[midway, right] {$\vb{w}_{nt}$};
    \end{tikzpicture}
    \caption{Illustration of subspace removal. We project a toxic sentence representation onto the toxic subspace. We remove this component from the original representation to create the non-toxic representation. Note that this modified representation is now orthogonal to the toxic subspace}
    \label{fig:my_label}
\end{figure}
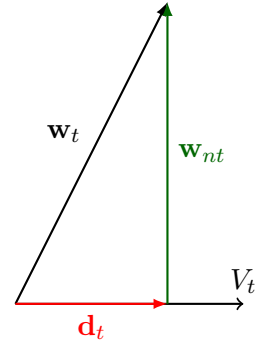

We adapt the approach developed by \cite{liang-etal-2020-towards} to remove the toxic subspace. Intuitively, for a given set of toxic sentence vectors $\vb{w}_t$, we subtract each toxic component $\vb{d}_t$ to produce the non-toxic representation.
\begin{equation} \label{eq:2}
    \vb{w}_{nt} = \vb{w}_{t} - \vb{d}_t
\end{equation}
However, when inferring new data, the value of $\vb{d}_t$ is unknown, and therefore we must approximate $\vb{d}_t$. 
We achieve this by projecting the sentence representations onto the lower dimensional toxic subspace $V_t$, then reconstructing the values in the original dimension. Thus, $\vb{d}_t \approx \langle \vb{w}_t, V_t \rangle V_t$. Substituting $\langle \vb{w}_t, V_t \rangle V_t$ into equation \ref{eq:2} yields a formula to remove an approximation of the toxic direction from toxic sentences. 
\begin{equation} \label{eq:1}
    \vb{\hat{\vb{w}}} = \vb{w}_t - \langle \vb{w}_t, V_t \rangle V_t
\end{equation}
In principle, $\vb{\hat{\vb{w}}}_i$ has no remaining bias, as it is orthogonal to $V_1$, and therefore has no bias components. We provide an intuitive illustration of the subspace removal in Figure \ref{eq:1}. Suppose that our toxic vector lies in some 2-D space, and we project the vector onto a 1-D toxic direction. The resulting difference between original vector and the projection yields the neutral vector.

\subsection{Selecting Principal Components}
We provide a heuristic to select the eigenvectors that explain the direction of toxicity. Since we perform principal component analysis on $\vb{d}_t \cup \vb{d}_{nt}$, the set of eigenvectors in $V$ accounts for both $\vb{d}_t$ and $\vb{d}_{nt}$. To isolate the subset of eigenvectors that account specifically for the toxic direction, we examine the principal component analysis reconstruction error.
\begin{equation*}
    \textit{Error} = \norm{\vb{D} - \langle \vb{D}, V \rangle V}_F
\end{equation*}
Here the set of difference vectors $\vb{D}$ has dimension $m\times d$, where $m$ is the number of observations and $d$ is the embedding dimension, and $V_t$ is a set of eigenvectors of dimension $n \times d$, where $n$ is the number of selected principal components selected. In practice however, per equation \ref{eq:1}, rather than projecting difference vectors onto the toxic subspace by performing $\langle \vb{D}, V \rangle V$, we instead project the sentence embeddings onto the toxic subspace by performing $\langle \vb{W}, V \rangle V$. Thus, we propose a more relevant measure of error. 
\begin{equation*}
    \textit{Error} = \norm{\vb{D} - \langle \vb{W}, V\rangle V}_F
\end{equation*}

We use our newly defined error to illustrate the importance of selecting only the eigenvectors that explain the toxic direction $\vb{d}_t$. Suppose to the contrary that we use the entire set of eigenvectors $V$. To minimize error, $\vb{D}$ must approximate the reconstructed values $\langle \vb{W}, V \rangle V$, which implies $\vb{d}_t \approx \langle \vb{w}_t, V \rangle V$ and $\vb{d}_{nt} \approx \langle \vb{w}_{nt}, V \rangle V$. However, when we rearrange $\vb{d}_{nt} \approx \langle \vb{w}_{nt}, V \rangle V$ to match equation \ref{eq:1}, we obtain $\vb{w}_t \approx \vb{w}_{nt} - \langle \vb{w}_{nt}, V \rangle V$, indicating the removal of the non-toxic direction from non-toxic representations to create toxic representations. Since we must strictly avoid the addition of any toxicity to our representations, we cannot include the eigenvectors that account for $\vb{d}_{nt}$.

To avoid this behavior, we select eigenvectors $V_t$ that fulfill the following criteria.
\begin{enumerate}
    \item $\min \norm{\vb{d}_t - \langle \vb{w}_t, V_t \rangle V_t}_F$
    \item $\max \norm{\vb{d}_{nt} - \langle \vb{w}_{nt}, V_t \rangle V_t}_F$
\end{enumerate}
In the first case, eigenvectors that can adequately transform a toxic representation $\vb{w}_t$ into a toxic direction $\vb{d}_t$ are desired. In the second case, eigenvectors that transform a non-toxic representation $\vb{w}_n$ into a non-toxic direction $\vb{d}_n$, facilitating the translation of non-toxic sentences into toxic sentences, are not desired. Note that, since the explained variance of eigenvectors produced by PCA can vary, the performance of this heuristic depends on the amount of explained variance associated with the eigenvector.

An alternate method of computing the toxic subspace $V_t$ is to perform subspace removal on only $\vb{d}_t$. While this approach does not need to filter eigenvectors, it nonetheless requires centering the features in $\vb{d}_t$ when performing principal component analysis, and consequently transforms the toxic directions. Thus, when approximating $\vb{d}_t$, we must transform the results back to the original center.

\begin{equation*}
    \vb{\hat{\vb{w}}}_{nt} = \vb{w}_t - \langle \vb{w}_t - \vb{\Bar{x}}_t, V_t \rangle V_t - \vb{\Bar{x}}_t
\end{equation*}
Here we define $\vb{\Bar{x}}_t$ as the feature means of the input vectors $\vb{d}_t$. Centering each feature introduces noise when inferring new data, as there is no guarantee that the means computed from the training data suit the new data. As a result we do not use this approach. 

\subsection{Debiased Toxicity Classifier}
Our context based word masking approach is contingent on a sentence-level toxicity binary classifier \textit{MaskerModel}. However, prior literature has shown that models of such nature associate protected groups with toxicity, and consequently identify non-toxic references to protected groups as toxic. We recognize the possibility that our classifier may make biased masking decisions, such as being predisposed to censor references to protected groups. To minimize the amount of biased masking decisions, we use the debiased toxicity classifier produced by \citep{46743} as the \textit{MaskerModel} in our context based masking algorithm. The model uses GloVe embeddings and is trained on debiased data based on the Wikipedia Toxicity Subtypes dataset. The training data is debiased by balancing references to protected groups in toxic contexts with references to protected groups in non-toxic contexts. 

\section{Experiments}
We aim to determine the performance of our approach in identifying and removing a subspace common to all pretrained BERT embeddings. To that end, we test our approach on several datasets and report the results below. Our results indicate that our approach both successfully removes toxicity while preserving semantic information. 

\subsection{Datasets}
We test our approach on the following datasets: 1) Wikipedia Toxicity Subtypes, 2) Civil Comments, and 3) Real Toxicity Prompts. From each dataset we construct a two-class sub-dataset to facilitate binary classification in downstream tasks. For each dataset we filter toxic examples using a binary toxicity classifier to reduce the number of observations close to the decision boundary. We keep only examples where the final softmax output exceeds $.8$ for the toxic label. 

\subsubsection{Wikipedia Toxicity Subtypes}
The Wikipedia Toxicity Subtypes is a collection of human annotated comments from Wikipedia Talks archives provided by \citep{10.1145/3038912.3052591}. Each comment has a label $0$ or $1$, where $0$ indicates a majority of annotators found the comment non-toxic, and $1$ indicates a majority of annotators found the comment toxic. 
\subsubsection{Civil Comments}
The Civil Comments dataset is a set of human annotated comments from the Civil Comments platform provided by \citep{DBLP:journals/corr/abs-1903-04561}. Each comment has a toxicity score on a scale from $0$ to $1$, representing the fraction of annotators that considered a comment toxic. We extract comments with a toxicity score greater than $.75$ and comments with a toxicity score exactly equal to $0$.
\subsubsection{Real Toxicity Prompts}
Real Toxicity Prompts is a dataset of natural text generation prompts created by \citep{RealTox} to study toxic text generation. Sentences are compiled from the Open Web Text corpus and automatically labelled with a toxicity score on a scale from $0$ to $1$ using the Perspective API, with $0$ being completely non-toxic and $1$ being completely toxic. We select only sentences with toxicity scores less than $.3$ and greater than $.7$ to increase polarity and reduce noise when performing downstream tasks. 

\subsection{Experimental Setup}

\subsubsection{Evaluation Metrics}
As outlined in our methodology, we measure the quality of our subspace using a modified PCA reconstruction error. We primarily use the error associated with reconstructing toxic direction vectors from a given sentence embedding, $\norm{\vb{d}_t - \langle \vb{w}_t, V_t \rangle V_t}_F$.

To determine the extent to which our approach removes toxicity, we calculate the number of toxic sentences before and after subspace removal. Let $\vb{X'_T}$ represent a set of sentences containing either toxic or non-toxic examples. Given \textit{Encoder}, a pretrained language model encoder such as BERT with parameters $\theta$, a binary classifier \textit{EvalModel} built on top of \textit{Encoder} that can identify toxic texts, and a toxic subspace $V_T$, we count the number of toxic inferences \textit{EvalModel} makes on $\vb{X'_T}$ after we modify the embeddings by removing $V_T$.

Maintaining semantic similarity before and after toxicity removal is another important constraint. Therefore, to determine whether neutralized representations bear similarity to the original sentence representation, we measure embedding cosine similarity before and after subspace removal. For each task we present the average cosine similarity achieved. 

Here we define these metrics in more detail. We define \textit{Toxic} as the number of toxic-labeled sentences that binary classifier $B$ correctly identifies as toxic and define \textit{Non-Toxic} as the number of neutral-labeled sentences that binary classifier $B$ erroneously identifies as toxic. We provide the classification accuracy as \textit{Acc} and two cosine similarity scores, \textit{Cos} and $\mathit{Cos_T}$. \textit{Cos} provides the average cosine similarity between original and modified sentence embeddings for all sentences, and $\mathit{Cos_T}$ provides the average cosine similarity between original and modified sentence embeddings for only toxic-labeled sentences. 

\subsubsection{Implementation Details}

For each dataset we evaluate the performance of toxicity removal. Per our evaluation metric, this requires the creation of a binary classifier \textit{EvalModel} for each dataset. Here we define binary classifier \textit{EvalModel} in more detail: Given a pretrained BERT encoder \textit{Encoder}, we add a fully connected linear layer over the pooled BERT encoding. When training, we freeze the training weights in \textit{Encoder} to ensure that we do not alter the latent space. Doing so would render the learned subspace $V_T$ invalid and make subspace removal impossible. Since \textit{EvalModel} is a fairly simple classifier, we use a learning rate of $.01$ to prevent overfit. For each dataset we set-aside a validation set of $1000$ toxic labeled sentences and $1000$ non-toxic labeled sentences. We undersample the remaining non-toxic sentences to balance the dataset, then train \textit{EvalModel} on the resulting data. We define the classifier trained on Civil Comments as \textit{CC}, the classifier trained on Wikipedia Toxicity Subtypes as \textit{WTS}, the classifier trained on Real Toxicity Prompts as \textit{RTP}.

Since our debiased \textit{MaskerModel} was trained on Wikipedia Toxicity Subtypes, we tested its generalization performance on other corpora. On the validation set for Civil Comments, \textit{MaskerModel} achieved an accuracy of $.902$, and on the validation set for Real Toxicity Prompts, it achieved an accuracy of $.908$, against a performance of $.952$ on Wikipedia Toxicity Subtypes.

We use each validation set to tune the set of eigenvectors selected out of the unfiltered collection of eigenvectors $V$. As stated previously, the performance of the eigenvector selection heuristic is closely related to explained variance. While our heuristic can reliably find the correct eigenvectors for the first several components, because explained variance decays, selecting subsequent less important components requires trial and error. To evaluate the performance of the best subspace, we test the subspace on out of distribution toxicity data. For instance, for the subspace derived from Wikipedia Toxicity Subtypes, we evaluate subspace removal performance on the validation sets for Civil Comments and Real Toxicity Prompts. 

Our experiments were computed using Google Colab's GPU backend (25 GB RAM, 147 GB Disk), with a random seed of 42. In all our evaluations, we perform principal component analysis and keep $32$ components. We then use our eigenvalue selection heuristic to select the top seven eigenvectors to represent the toxic subspace. We explore why we select seven in the Eigenvector Selection section below. 

\subsection{Toxicity Removal}

\begin{table}[tp]
\center
\begin{tabular}{@{}cc|ccc@{}}
\toprule
\textbf{Dataset} & \textbf{Removal} & \textbf{Tox} & \textbf{Non-Tox} & \textbf{Acc.}  \\ \midrule
\textbf{Civil}  &  \textbf{Yes} & 0 & 0 & 0.500 \\
\textbf{} & \textbf{No} & 834 & 300 & 0.767 \\ \midrule
\textbf{Wiki} & \textbf{Yes} & 14 & 6 & 0.557 \\
\textbf{} & \textbf{No} & 867 & 134 & 0.867 \\ \midrule
\textbf{Real} & \textbf{Yes} & 98 & 22 & 0.538 \\
\textbf{} & \textbf{No} & 890 & 216 & 0.837 \\ \bottomrule
\end{tabular}
\caption{\label{table:1}Validation results for toxicity removal. For each dataset we present the size of the set of toxic directions fed into the principal component analysis, as well as results for toxic subspace removal and no toxic subspace removal. We define \textit{Tox} as the number of toxic predictions of toxic sentences after subspace removal, \textit{Non-Tox} as the number of toxic predictions of non-toxic sentences, and \textit{Acc.} as the classification accuracy.}
\end{table}
We present our validation results in Table \ref{table:1}, and evaluation results in Table \ref{table:2}. In Table \ref{table:1} we report the performance of each classifier \textit{CC}, \textit{WTS}, and \textit{RTP} without any subspace removal as a baseline. We find a drastic decrease in toxic predictions compared to this baseline when we remove the toxic subspace, among all datasets we used. 

Performance on the test sets further demonstrates the success of our approach. In Table \ref{table:2}, for each dataset, removing the derived subspace can effectively reduce the number of toxic predictions. For subspaces derived from Civil Comments and Wikipedia Toxicity Subtypes, the number of toxic predictions after subspace removal drops to less than $5\%$ the original amount of toxic texts, and for Real Toxicity Prompts, the number of toxic predictions decreases to $20\%$. Additionally, we show that the classification accuracy on each evaluation set of $1000$ toxic and non-toxic sentences has decreased to approximately $.5$. 

Furthermore, effectiveness in removing the toxic subspace does not yield a large adverse affect on the similarity before and after subspace removal. For the most part, we report average cosine similarities ranging from $.7$ to $.8$. These results show that a subspace derived from one toxic corpus generalizes well to other toxic corpora. This suggests first that an underlying lower dimensional toxic subspace exists in pretrained language model embeddings, and second that we can separate toxicity from a sentence without significantly altering its meaning. 

\begin{table*}[pt]
\centering
    \begin{tabular}{@{}cc|ccccc@{}}
    \toprule
    \textbf{Train} & \textbf{Test}  & \textbf{Tox} & \textbf{Non-Tox} & \textbf{Cos} & \textbf{Cos}$_t$ & \textbf{Acc.} \\ \midrule
    \textbf{Civil}    & \textbf{Real}  & 28           & 3                & 0.806        & 0.812         & 0.513         \\
    \textbf{}         & \textbf{Wiki}  & 1            & 0                & 0.801        & 0.812         & 0.501         \\ \midrule
    \textbf{Wiki}     & \textbf{Civil} & 3            & 0                & 0.734        & 0.743         & 0.502         \\
    \textbf{}         & \textbf{Real}  & 37           & 4                & 0.767        & 0.772         & 0.516         \\ \midrule
    \textbf{Real}     & \textbf{Civil} & 42           & 13               & 0.681        & 0.689         & 0.515         \\
    \textbf{}         & \textbf{Wiki}  & 170          & 23               & 0.712        & 0.72          & 0.574         \\ \bottomrule
    \end{tabular}
\caption{\label{table:2}Evaluation results for toxicity removal. We evaluate performance of each model on out of distribution toxic samples. We define \textit{Cos} as the similarity before and after subspace removal for all sentences and \textit{Cos}$_t$ as the similarity for only toxic sentences. For all other metrics we use the same definitions as in Table \ref{table:1}. }
\end{table*}

\begin{table}[tp]
\begin{tabular}{@{}c|ccc@{}}
\toprule
 \textbf{Dataset} & \textbf{$|\vb{W}|$} & \textbf{$V_t$ scaled err.} & \textbf{$V$ scaled err.} \\ \midrule
\textbf{Civil} & 10682 & 1.488 & 3.219 \\
\textbf{Wiki} & 6462 & 1.714 & 3.254 \\
\textbf{Real} & 4992 & 2.346 & 4.173 \\ \bottomrule
\end{tabular}
\caption{\label{table:3} Scaled PCA reconstruction error. For each dataset we report the size of the PCA input data $\vb{W}$. We report the errors when projecting sentence vectors onto the toxic subspace $V_t$ and the larger unfiltered subspace $V$. We scale error values by the Frobenius norm of the toxic direction vectors $\vb{d}_t$.}
\end{table}

Our results also show that our approach can remove toxicity from toxic sentences without adding toxicity to non-toxic sentences. Moreover, for all datasets, the number of incorrectly classified non-toxic sentences decreases before and after subspace removal.

We present the error for each subspace in Table \ref{table:3}. Because the magnitude of our error is affected by the dataset size, we scale our results by the Frobenius norm of the set of toxic direction vectors $\vb{d}_t$, providing an error that is relative to the norm of the direction vectors. We find that the larger the PCA input, the better the subspace performs at both removing toxicity and preserving content. Additionally, these results would suggest that the subspace derived from Real Toxicity Prompts performs worst at toxicity removal, and that the subspace derived from Civil Comments performs best at toxicity removal. Indeed, the values from Table \ref{table:2} confirm these assumptions. We thus demonstrate a correlation between our error values and our experimental results, highlighting the validity of our approach. 

\subsection{Parallel Corpus Evaluation}

\begin{table}[ht]
\begin{tabularx}{\linewidth}{@{}l|X@{}}
\toprule
 & \textbf{Sentence} \\ \midrule
Toxic &  you can't be this stupid oh right this is oregon \\
\textit{Non-Toxic} &  \textit{you can't be this [MASK] [MASK] right this is oregon} \\ \midrule
Toxic &  you're a moron the [MASK] paint a different picture we have needs all over the damn place you want to [MASK] your head in the sand \\
\textit{Non-Toxic} & \textit{you're a [MASK] the [MASK] paint a different picture we have needs all over the [MASK] place you want to [MASK] your head in the sand} \\ \midrule
Toxic & like father like scum \\
\textit{Non-Toxic} & like father like [MASK] \\ \midrule
Toxic &  your defense of trump is simply pathetic really pathetic \\
\textit{Non-Toxic} & your defense of trump is simply [MASK] really pathetic \\ \bottomrule
\end{tabularx}
\caption{\label{table:check} Four randomly selected examples. We replace out of vocabulary tokens in the original toxic sentence with the ``[MASK]" token.}
\end{table}

We provide a brief evaluation of our parallel corpus creation algorithm. We randomly select a sample of four toxic and corresponding non-toxic pairs from Civil Comments to illustrate the effect of our token masking algorithm which we present in Table \ref{table:check}. Borrowing the definition of toxicity defined by \citeauthor{DBLP:journals/corr/abs-1903-04561}, in which toxicity is any rude language that would cause an individual to leave a conversation, we find that our method of token masking generally masks the tokens that contribute to toxicity. No approach is perfect, however, and our approach occasionally misses toxic tokens, as in the last case in Table \ref{table:check}. Nevertheless, since principal component analysis inherently performs noise reduction, our approach can tolerate a moderate degree of inaccuracy from our token masking approach and still produce a valid toxic subspace.

While masking toxic tokens has the potential to alter significantly the meaning of the original sentence, we show that this is not the case. We provide average cosine similarity values between each toxic and non-toxic sentence pair along with the standard deviation, computed \textit{before} cleaning data to remove observations with low similarity scores. On Civil Comments, the average similarity between original and masked sentences was $0.948$, with a standard deviation of $0.064$. On Wikipedia Toxicity Subtypes, the average similarity between original and masked sentences was $0.937$ with a standard deviation of $0.073$. Finally, on Real Toxicity Prompts, the mean similarity and standard deviation was $0.968$ and $0.050$ respectively. These values indicate that our token masking approach generally does not significantly alter the meaning of the original sentence.

\subsection{Evidence of Subspace}

\begin{figure} [ht]
    \centering
    \includegraphics[scale=.5]{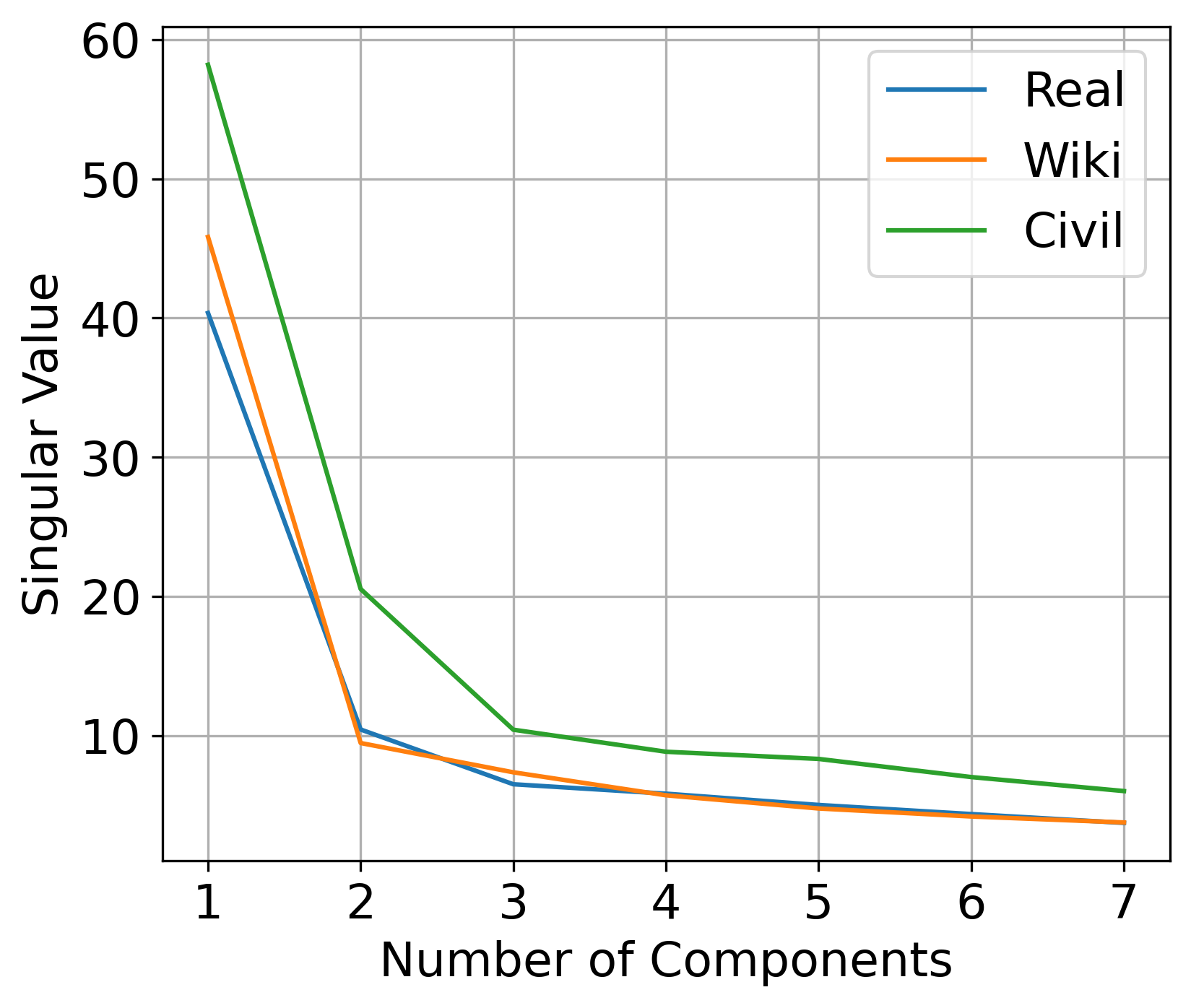}
    \caption{Singular values for each of the first seven principal components for the \textit{toxic subspace}. We show that most of the variance in our toxic subspace is concentrated within the first several principal components.}
    \label{fig:ev}
\end{figure}

To illustrate the existence of a toxic subspace, we present the singular values for each principal component of the toxic subspace we have created. Having the variance in toxic direction $\vb{d}_t$ concentrated in the first several principal components strongly indicates that the principal component analysis has successfully found the subspace. We plot the singular values obtained by performing principal component analysis on Real Toxicity Prompts, Wikipedia Toxicity Subtypes, and Civil Comments in Figure \ref{fig:ev}. For each dataset, we observe that most of the variance is concentrated within the top several principal components. Past the initial principal components, the singular values begin to converge. Thus, we find evidence indicating the existence of the toxic subspace.

\subsection{Eigenvector Selection}

\begin{figure*}[ht]
    \centering
    \includegraphics[scale=.5]{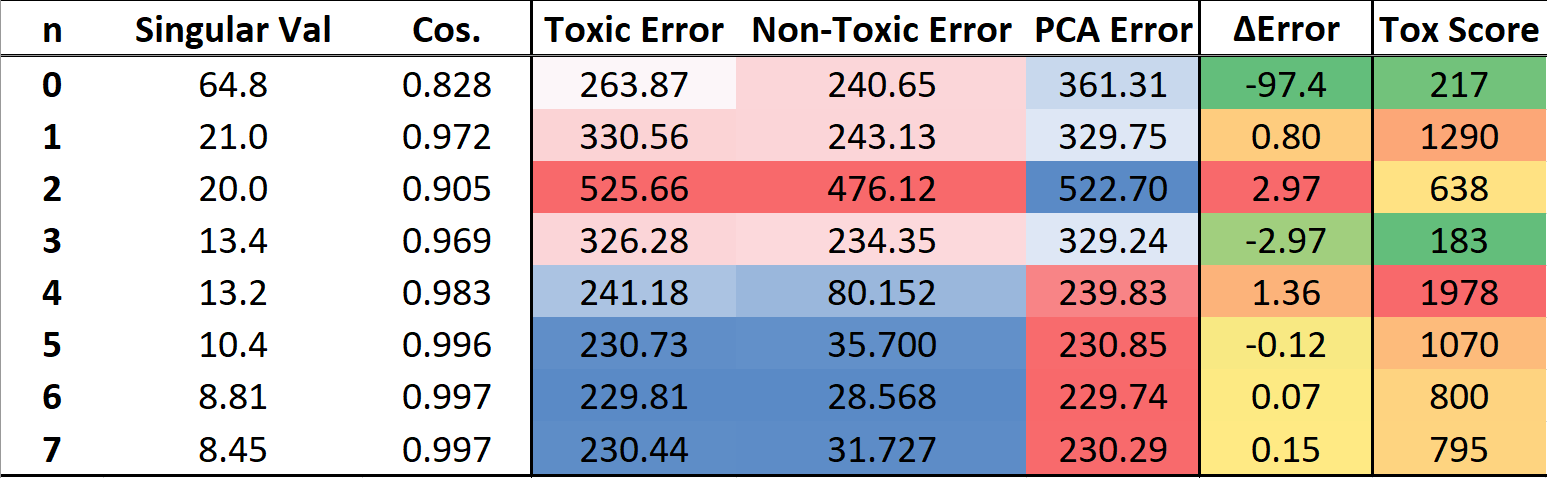}
    \caption{Analysis of eigenvector selection heuristics. We perform subspace removal with the $n$th eigenvector, one eigenvector at a time. We define \textit{Toxic Error} as the error associated with producing a non-toxic sentence from a toxic sentences, \textit{Non-Toxic Error} as the error associated with keeping non-toxic sentence non-toxic after subspace removal, \textit{PCA Error} as the error associated with producing a toxic sentence from a non-toxic sentence, $\Delta$\textit{Error} as \textit{Toxic Error} - \textit{PCA Error}, and \textit{Tox Score} as the number of toxic predictions observed after subspace removal. We color code the data to highlight trends present. Blue and green indicate desired values, while red indicates undesired values.}
    \label{fig:analysis}
\end{figure*}

We perform an analysis of the eigenvector selection heuristic outlined previously.
For each eigenvector in the unfiltered set of eigenvectors $V$, we compute and report relevant values for our heuristic in Figure \ref{fig:analysis}. All values reported in Figure \ref{fig:analysis} are derived from the Wikipedia Toxicity Subtypes dataset. We define \textit{Toxic Error} $\norm{\vb{d}_t - \langle \vb{w}_t, V \rangle V}_F$ as the error when approximating the toxic direction $\vb{d}_t$ using toxic sentence representations $\vb{w}_t$. We define \textit{PCA Error} $\norm{\vb{d}_n - \langle \vb{w}_n V \rangle V}_F$ as the error when approximating the neutral direction $\vb{d}_n$ using toxic sentence representations $\vb{w}_n$. Lastly we define a third error measurement \textit{Non-Toxic Error} derived from the desired outcome when removing toxicity from neutral sentences: $\vb{w}_n \approx \vb{w}_n - \langle \vb{w}_n, V \rangle V$. This equation relates that subspace removal should not alter the meaning of non-toxic sentences. Rearranging the equation yields $\norm{\langle \vb{w}_n, V \rangle V}_F$. We report the difference between Toxic Error and PCA Error as $\Delta \textit{Error}$ and the number of toxic predictions after subspace removal as \textit{Tox Score}. 

In Figure \ref{fig:analysis}, larger negative values for $\Delta$ Error tend to correlate with improved Tox Score. This observation supports our heuristic for selecting eigenvectors, where we aim to minimize Toxic Error and maximize PCA Error. Additionally, based on the results in Figure \ref{fig:analysis}, we do not include Non-Toxic Error as a heuristic for selecting eigenvectors. We find that Non-Toxic Error is very well correlated with Toxic Error, and thus, while minimizing Non-Toxic Error is important, its function as a heuristic is made redundant by Toxic Error. In effect, minimizing the amount of error in transforming toxic sentences into non-toxic sentences also tends to minimize the error in keeping non-toxic sentences non-toxic after toxicity removal.

We note that the contribution of the $n$-th principal component decreases as $n$ increases. In Figure \ref{fig:analysis}, as $n$ increases, singular value decreases, and the size of the changes made in the latent space (as indicated by the cosine similarity before and after subspace removal) converges to a minimal value. We determine experimentally that after $7$ components, the benefit of additional principal components is negligible. Thus, we limit the size of our toxic subspace to the first $7$ principal components.

At $7$ principal components, the toxic subspace derived from Wikipedia Toxicity Subtypes was found using only the eigenvalue selection heuristic and did not need any fine-tuning. However, this was not true for the other two datasets. The Civil Comments subspace required fine-tuning after selecting the third eigenvector from the set of unfiltered eigenvectors $V$, and the Real Toxicity Prompts subspace required fine-tuning after selecting the second eigenvector.

\section{Conclusion}

In this paper we create a method to identify an underlying toxic subspace within pretrained language model embeddings. Additionally, we propose a parallel corpus creation method by masking toxic words and phrases. Through our experiments, we show that the subspace we find drastically decreases the amount of toxicity and may even preserve semantic meaning. We test our approach on several toxicity datasets and determine that a toxic subspace found in one corpus explains toxicity in other corpora. This evidence strongly indicates that we have created a method to find an underlying toxic subspace within pretrained language models.
The existence of this toxic subspace suggests that toxic features follow some underlying pattern, and further suggests that we can use this subspace to prevent toxic features in generated text.

\bibliographystyle{acl_natbib}
\bibliography{biblio,acl2021}


\end{document}